\documentclass{vgtc}                          % final (conference style)
\graphicspath{{figures/}{pictures/}{images/}{./}} % where to search for the images

\usepackage{times}                     % we use Times as the main font
\usepackage{tabu}                      % only used for the table example
\usepackage{booktabs}                  % only used for the table example
\usepackage{lipsum}                    % used to generate placeholder text
\usepackage{mwe}                       % used to generate placeholder figures

\usepackage{mathptmx}                  % use matching math font

\usepackage{makecell}
\usepackage{amsmath}
\usepackage{bbding}
\usepackage{multirow}

\usepackage{caption}
\usepackage{graphicx}
\usepackage{amsfonts}

\onlineid{3664}

\vgtccategory{Research}

\vgtcinsertpkg

\title{HumanCoser: Layered 3D Human Generation via Semantic-Aware Diffusion Model}

\author{
    Yi Wang$^{1\,2}$\setcounter{footnote}{1}\thanks{Equal contribution.}, 
    Jian Ma$^{1 \dagger}$, 
    Ruizhi Shao$^{3}$, 
    Qiao Feng$^{1}$, 
    Yu-Kun Lai$^{4}$, 
    Kun Li$^{1}$\setcounter{footnote}{0}\thanks{Corresponding author.} \\\\
    $^{1}$Tianjin University, China \enspace  $^{2}$Changzhou Institute of Technology, China \\
    $^{3}$Tsinghua University, China \enspace   $^{4}$Cardiff University, British \\
}

\teaser{
  \centering
  \includegraphics[width=\linewidth]{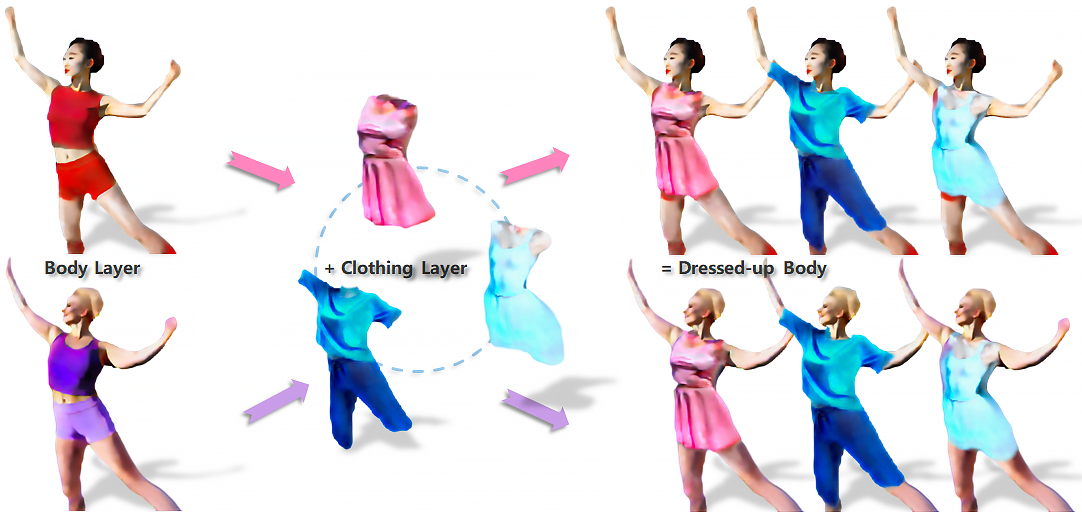}
  \caption{Our method can generate layered 3D humans guided by text prompts, which are physically-decoupled and structurally consistent. This allows our generated clothing to be reused, exchanging between digital avatars with different identities.}
  \label{fig:A1}
}

\abstract{
   This paper aims to generate physically-layered 3D humans from text prompts. Existing methods either generate 3D clothed humans as a whole or support only tight and simple clothing generation, which limits their applications to virtual try-on and part-level editing. To achieve physically-layered 3D human generation with reusable and complex clothing, we propose a novel layer-wise dressed human representation based on a physically-decoupled diffusion model. Specifically, to achieve layer-wise clothing generation, we propose a dual-representation decoupling framework for generating clothing decoupled from the human body, in conjunction with an innovative multi-layer fusion volume rendering method. To match the clothing with different body shapes, we propose an SMPL-driven implicit field deformation network that enables the free transfer and reuse of clothing. Extensive experiments demonstrate that our approach not only achieves state-of-the-art layered 3D human generation with complex clothing but also supports virtual try-on and layered human animation. More results and the code can be found on our project page at \url{https://cic.tju.edu.cn/faculty/likun/projects/HumanCoser}.
} % end of abstract


\begin{document}


\firstsection{Introduction}

\maketitle

The generation of 3D {humans} with changeable clothing plays an important role in movies, games and AR/VR. 
Existing methods~\cite{a1,a2,a3,a4,a9,a10} only produce a unified surface encompassing both the body and clothing, leading to a body-clothing coupling. 
% As a result, these methods struggle to edit or change the dressed human's clothing and body separately in a layer-wise manner, which restricts their ability to achieve detailed customization and accurate adjustments for virtual try-ons, animated character design, and personalized avatar creation.
This limits their ability to edit clothing and body separately, restricting detailed customization and accurate adjustments for virtual try-ons, animated character design, and personalized avatar creation.
% However, these coupled representation hardly edit or change the dressed human's clothing and body separately in layer-wise manner.
In this paper, we aim to generate high-fidelity layered 3D human which can be edited and exchanged for clothing via representation-decoupling, as shown in \cref{fig:A1}.

Recently, owing to the high-quality image synthesis capability of pre-trained {diffusion models}~\cite{a21}, methods~\cite{a9,a10,a11} introduce a novel Score Distillation Sampling ($\mathrm{SDS}$) strategy~\cite{a21} to self-supervise the 3D human generation process. However, these methods ignore the diversity and self-occlusion of human shapes, which leads to inconsistencies in generated human structures. 
Furthermore, most data-driven 3D avatar generation methods~\cite{a12,a13,a42,a43,a44,a45} generate 3D clothed humans in a coupled manner, {and as a result,} clothing cannot be exchanged between arbitrary bodies. Overall, the above methods fail to ensure structural consistency of the human body and lack the capability to generate and edit bodies and clothes in a layered and flexible manner.

This paper introduces HumanCoser, a novel framework based on a physically-decoupled diffusion model. It aims to generate representation-decoupling and animatable 3D dressed humans with consistent body structure in a layer-wise manner, guided by text. To achieve accurate layer-wise clothing representation, we propose a dual-representation decoupling framework designed to generate clothing independent from the human body. 
This framework is complemented by an innovative multi-layer fusion volume rendering method. HumanCoser, thus, effectively generates multi-layer clothing consistent with the text prompts. Moreover, to ensure accurate geometric alignment between decoupled clothing and the body, we present a 3D implicit deformation field leveraging SMPL [47] as a clothing proxy for matching clothing with the body. Furthermore, to enhance details, we introduce a normal prediction network for smooth normals, combined with optimized spherical harmonic (SH) lighting. Hence, the proposed HumanCoser can generate reusable and intricate multi-layered dressed 3D humans that can be edited and changed separately as shown in \cref{fig:A1}.

Our main contributions are summarized as follows:
\begin{itemize}
\item 
We propose a layered 3D human generation framework with a multi-layer representation decoupling method. To {our} best knowledge, this is the first work that can make the 3D dressed human truly decoupled physically and support layered generation and editing for 3D dressed human. We also introduce a decoupled shape prior to generating structurally consistent {3D} content.

\item 
We propose a dual-representation decoupling strategy to improve the semantic consistency of generated clothing, combined with an innovative multi-layer fusion volumetric rendering approach. The strategy not only improves the semantic consistency of the clothing but {is} also {generalizable} to the enhancement of 3D semantics for other wearable outfits of humans.
\item 
% \textcolor{red} {
We propose a 3D implicit deformation method based on SMPL vertex prediction to achieve the geometric matching of human bodies and clothing in {an} implicit manner so that the clothing can be transferred between different human {subjects}.
\end{itemize}

\section{Related Work}
\label{sec:related}
\noindent \textbf{Text-guided 3D Content Generation.} CLIP-Forge~\cite{a18} and Dream-Field~\cite{a19} optimize {Neural Radiance Fields (NeRFs)} 
to generate 3D shapes by aligning the embedding of the generated image with the text embedding in the space of the image-text model CLIP. CLIP-mesh~\cite{a49}
also uses CLIP to optimize {meshes} to represent 3D shapes. {However,} by directly generating images aligned with text in CLIP space, it is not possible to generate highly realistic images. Recently, diffusion modeling~\cite{a20,a21,a22} has {seen rapid growth} due to its excellent performance in synthesizing high-quality images. DreamFusion~\cite{a7} proposes Score Distillation Sampling ($\mathrm{SDS}$) based on a pre-trained diffusion model~\cite{a21} to optimize trainable NeRFs. Magic3D~\cite{a6} uses a 2-stage training strategy to bootstrap 3D texture networks to optimize 3D content generation. Although the above {diffusion}-based 3D generation models have some 3D generation capability, the generation of 3D human is a challenge for the above methods due to the complexity of their shapes and the diversity of their poses.

\noindent \textbf{Text-guided 3D Human Generation.} AvatarCLIP~\cite{a23} initializes the 3D human body shape via a VAE {(Variational Autoencoder)} encoder and then performs geometric shaping and texture generation guided by an image-text model~\cite{a24}. However, since the method focuses on shaping localized structures, it lacks in the generation of global structures such as skirts, long hair and loose clothing. In addition, Latent-NeRF~\cite{a11} and TADA~\cite{a46} both utilize pre-trained text-to-image diffusion models for 3D avatar generation work. In particular, Latent-NeRF~\cite{a11} employs a Sketch-Shape to constrain the generation of the diffusion model, but the results of this method lack {details} due to the lack of {optimization} of normals and illumination. %While 
TADA~\cite{a46} is limited by the representation ability of the confined mesh, {and thus} cannot represent non-convex structures {or} transparent materials well. {Neither} of the above methods can generate 3D avatars with layer-wise bodies and clothing. 
Dreamhuman~\cite{a53} produces animatable coupled avatar based on text and human posture. It combines 3D human prior to generate and re-pose the generated results, but it cannot arbitrarily adjust and replace the clothes of humans without retaining human identity. Avatarcraft~\cite{a50} transforms text into a 3D avatar, using a diffusion model to stylize geometry and texture, while shape and pose are controlled by a parametric human model. Avatarcraft uses a bare neural human avatar as a template. Given a text prompts, Avatarcraft uses the diffusion model to guide the creation of the avatar by updating the template so that the geometry and texture are consistent with the text. Although Avatarcraft updates the avatar with new pose and shape parameters without training, the generated avatar hardly shows details such as loose clothing and fluffy hair. 
Dreamwaltz~\cite{a52}  generates 3D digital avatars from text prompts, leveraging prior knowledge of human body shapes and poses, and facilitating animation and interactive compositions between avatars, objects, and scenes. It learns the distribution of human animations through prior knowledge of human actions, enabling the generation of plausible human animations. However, Dreamwaltz's learnable human action deformation module lacks generalizability for generating multi-layered humans, thereby hindering capabilities such as dress-up and clothing editing.
DreamAvatar~\cite{a51} uses SMPL for shape guidance and introduces a dual-observation space design to optimize shape and pose jointly. It addresses the ``Janus'' problem and enhances facial details. However, it fails to fully consider human body occlusion information, and it also couples clothing and human body generation. Distinct from non-layered methods, HumanLiff~\cite{a48} generates human body based on the diffusion model in a layer-by-layer manner. However, the features depend on the tri-plane features of the previous layer. This coupling of features among layers impedes the separate editing and reuse of each layer. 

In summary, existing methods either generate 3D dressed humans as a whole or support only tight and simple clothing generation, which limits their applications to virtual try-on and part-level editing. In contrast, our method can generate reusable and intricate multi-layered 3D dressed humans that can be edited and changed separately. It achieves realistic body and clothing generation by predicting normals and employing improved spherical harmonic lighting. Moreover, we ensure the semantic consistency of the generated clothing through an optimized dual-representation decoupling framework. The layered clothing can be seamlessly transferred between different shapes of human bodies using an implicit deformation network based on SMPL. We summarize the main differences between our work and related work in \cref{tab:compare}.

\begin{table}[t]
 \caption{Comparison of 3D human generation methods, including layered generation, geometric complexity, clothing transfer and clothing reusability.}
  \scriptsize%
  \centering%
  \scalebox{0.8}{
  \begin{tabular}{|c|cccc|}
    \toprule
     Method & Multilayer & \makecell[c]{Geometry\\(non-skin tight)}& Clothing Transfer & Reusability\\
    \midrule
    AvatarCLIP~\cite{a23} &  \XSolidBrush & \XSolidBrush & \XSolidBrush  & \XSolidBrush\\
    TADA~\cite{a46} & \XSolidBrush & \XSolidBrush & \XSolidBrush  & \XSolidBrush\\
    Latent-NeRF~\cite{a11} & \XSolidBrush & \Checkmark & \XSolidBrush  & \XSolidBrush\\
    HumanLiff~\cite{a48} & \Checkmark & \Checkmark & \XSolidBrush  & \XSolidBrush\\
    Ours & \Checkmark & \Checkmark & \Checkmark  & \Checkmark\\
    \bottomrule
  \end{tabular}}
  \label{tab:compare}
%\vspace{-0.4cm}
\end{table}
% \vspace*{-15pt}

\section{Method}
\label{sec:method}

\subsection{Overview}
\label{sec:4.0}

\begin{figure*}[t]
  \centering
   \includegraphics[width=0.95\linewidth]{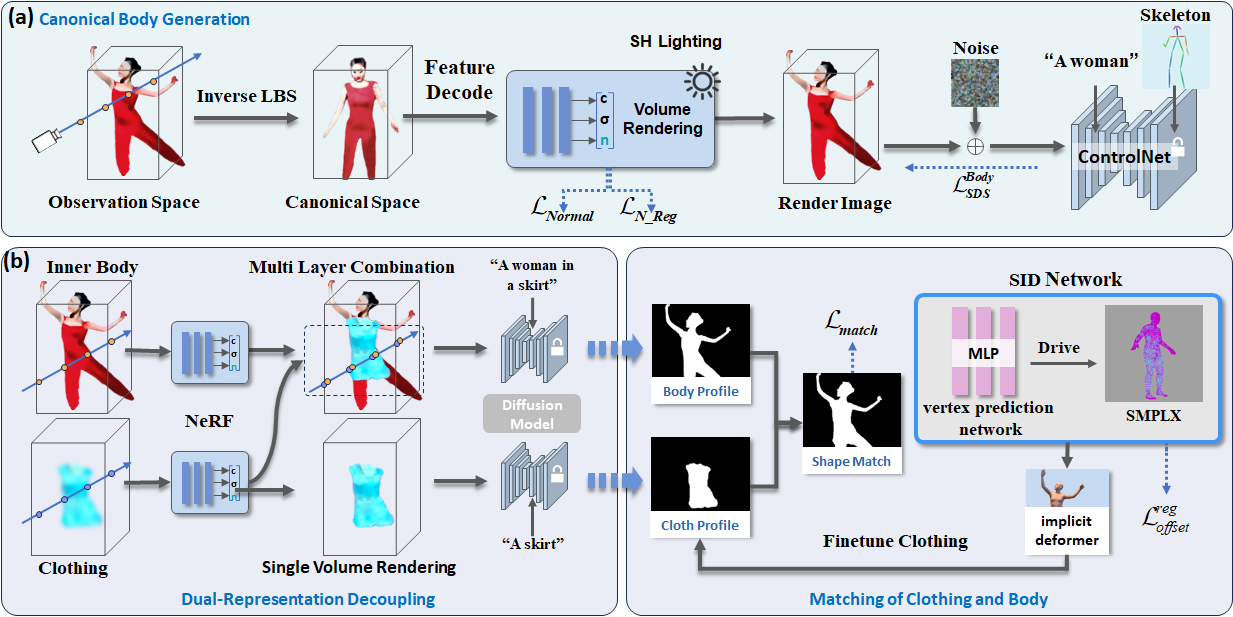}
   %\vspace{-0.2cm}
   \caption{Illustration of our framework for generating the clothes and body of a dressed human in a layered manner. (a) shows the generation of the minimized body, and (b) shows the layered generation of clothing and the matching of clothing with the body.}
   \label{fig:pipline}
   \vspace{-0.3cm}
\end{figure*}

The proposed HumanCoser is a two-stage method to generate realistic 3D humans with consistent body structures guided by text in a layer-wise manner. The first stage (a) is to generate a minimized human body, and the second stage (b) performs decoupled generation of clothing and matches the clothing with the human body. Specifically, as \cref{fig:pipline} shown, stage (a) consists of a NeRF and ControlNet with SMPL skeleton conditions as inputs and generates a minimized human body in canonical space (\cref{sec:3.2}). In addition, stage (b) consists of a dual-representation decoupling network (\cref{sec:3.3}) and an implicit deformation network driven by SMPL (\cref{sec:3.4}). In which, the dual-representation decoupling network generates dis-entangled clothing on the basis of the minimized human body combined with a multi-layer fusion rendering method. Finally, the decoupled clothing is matched with the human body through the deformation network mentioned above.

\subsection{Canonical Body Generation}
\label{sec:3.2}
In order to obtain the minimized human body, we adopt ControlNet with SMPL as conditional input and generate human body in canonical space, as shown in \cref{fig:pipline}. 
We first use NeRF as a representation of layered humans. The inner body and each layer of clothing are represented by a separate network as follows:
\begin{equation}
\begin{aligned}
F_\theta(\gamma(\mathbf{x}))=(\sigma, c),
\end{aligned}
\end{equation}
where $\gamma(\mathbf{.})$ is the frequency encoder. We render the scene using the volume rendering equation $F_\theta(.)$~\cite{a8}. $\sigma$ and $c$ denote the density and color predicted by each sampling point $\mathbf{x}$. The color of each layer is predicted as follows:
\begin{equation}
\begin{aligned}
&C(\mathbf{r})=\sum_i w_i {c}_i, \\
&w_i=\alpha_i \prod_{j<i}\left(1-\alpha_j\right),
\end{aligned}
\label{eq:eq2}
\end{equation}
where $\alpha_i=1-\exp \left(-\sigma_i\left\|\mathbf{x}_{i}-\mathbf{x}_{i+1}\right\|\right)$ and $\left\|\mathbf{x}_{i}-\mathbf{x}_{i+1}\right\|$ is the interval between sample ${i}$ and ${i}+1$. {$w_i$} is the weight of the {$i^{th}$} sampling point~\cite{a8}. {$c_i$} and $\sigma_i$ is the predicted color and density of the {$i^{th}$} sampling point~\cite{a8}.

\noindent 

\textbf{Multi-Layer Fusion Rendering.} 
In order to fuse the layered human body and clothing for rendering, we proposed a multi-layer composite rendering method based on the density and weight of each sampling point according to (\cref{eq:eq2}), which is defined as follows:
\begin{equation}
C^{\prime}(\mathbf{r})=\sum_{i=1}^N w_i^j c_i^j, w_i^j=\max \left(w_i^1 \cdots w_i^n\right) ,
\label{eq:eq3}
\end{equation}
where $C^{\prime}(\mathbf{r})$ is the rendering formula (\cref{eq:eq2}), $w_i^j$ is the weight with the highest density in the $n$-layer component, and $c_i^j$ is the corresponding color. In addition, in order to make the generated surface normals {smoother}, we calculate the normal loss between the predicted normal $\mathbf{n}^{\prime}$ and the surface normal $\mathbf{n}$:

\begin{equation}
\begin{aligned}
\mathcal{L}_{n}=\sum_i w_i\left\|\mathbf{n}^{\prime}_{i}-\mathbf{n}_{i}\right\|,
\end{aligned}
\label{eq:eq4}
\end{equation}
where {$w_i$} is the weight of the {$i^{th}$} sampling point, which follows the definition of \cref{eq:eq2}. Moreover, in order to regularize the normal and reduce redundant semantically generated artefacts, a loss of regular constraints is added:
\begin{equation}
\begin{aligned}
\mathcal{L}_n^{r e g}=\sum_i w_i\left(1-\sum \mathbf{n}^{\prime}_{i} \cdot \mathbf{n}_{i}\right),
\end{aligned}
\end{equation}
where $w_i$ is the weight of the {$i^{th}$} sampling point. Definitions of $\mathbf{n}^{\prime}_{i}$ and $\mathbf{n}_{i}$ refer to \cref{eq:eq4}. 
Furthermore, we introduce SDS loss to optimize 3D models of the body and clothing:
\begin{equation}
\nabla_{\boldsymbol{\theta}} \mathcal{L}_{\mathrm{SDS}}(\phi, \mathbf{x})=\mathbb{E}_{t, \boldsymbol{\epsilon}}\left[\mathbf{w}(t)\left(\boldsymbol{\epsilon}_\phi\left(\mathbf{x}_t ; y, t, \mathbf{c}\right)-\boldsymbol{\epsilon}\right) \frac{\partial \mathbf{x}}{\partial \boldsymbol{\theta}}\right]
\end{equation}
where $\mathbf{w}(t)$ represents a weight function dependent on the time step $t$, $\mathbf{x}$ is the noisy image, and $y$ is the input text prompt. The noise injected by $\epsilon$ is added to the rendered image $\mathbf{x}$. To maintain the consistency of the human body structure, we input the SMPL skeleton $\mathbf{c}$ as the conditional image.
Therefore, the overall loss of the canonical body generation is as follows:
\begin{equation}
\mathcal{L}_{body}=\lambda_{\mathrm{SDS}}^{\mathit{body}}  \mathcal{L}_{\mathrm{SDS}}^{\mathit{body}}+\!\lambda_n \mathcal{L}_n+\!\lambda_n^{\mathit{reg}} \mathcal{L}_n^{\mathit{reg}},
\end{equation}
where $\mathcal{L}_\mathrm{SDS}^{body}=\mathcal{L}_\mathrm{SDS}\left(\mathbf{x}^b; y^b, \mathbf{c}^b\right)$, $\mathbf{x}^b$ is the supervised body image, $ y^b$ is the prompt of body, and $\mathbf{c}^b$ is the input condition of SMPL skeleton.
$\lambda_{\mathrm{SDS}}^{\mathit{body}}$, $\!\lambda_n$, $\!\lambda_n^{\mathit{reg}}$ are the weights attributed to each loss.
More details of \cref{sec:3.2} are provided in the supplementary material.

%\vspace{-0.2cm}
\subsection{Dual-Representation Decoupling}
\label{sec:3.3}
In order to accurately obtain the shape of clothing, we introduce a dual-representation decoupling framework (DRD) to eliminate the parts that are inconsistent with the semantics of clothing.

%hero

\noindent
\textbf{Decoupling Clothing Representation.} As illustrated in \cref{fig:pipline}(b), the DRD model consists of a multi-layer component composition network and a clothing generation network. During the training of the clothing component at the $N^{th}$ layer, we take the density of the sampling point with the largest weight in the first $N-1$ layers as the combination density of the first $(N-1)^{th}$ layers. This combined density is defined as follows:
\begin{equation}
\underset{\delta}{\operatorname{argmax}}(w(\delta)), \delta \in\left\{\delta_1 \ldots \delta_{N-1}\right\},
\end{equation}
where $w(\mathbf{.})$ is defined in \cref{eq:eq2}. The final combination weight $w(\delta)$ is then calculated based on the combined density $\delta$. We use the following loss to constrain the density of the overlapping parts of the $N^{th}$ clothing and the first $N-1$ components:
\begin{equation}
\begin{aligned}
& \mathcal{L}_{{reg\_ds}}=\left\|w\left(\delta_c^N\right)\right\|_2, \\
& \left\{\delta_c^N \mid w\left(\delta_c^N\right)>\lambda \cup \delta_c^N<\delta_{b c}\right\},
\end{aligned}
\end{equation}
where $w(.)$ is \cref{eq:eq2} which only considers the input of density, $\delta_c^N$ is the density of the $N^{th}$ clothing component, $\delta_{b c}$ is the combination density of the first $N-1$ components, and $\lambda$ is the defined threshold. Finally, as shown in \cref{fig:pipline}(b), we perform a composite rendering of the $N^{th}$ clothing component and the first $N-1$ components based on the \cref{eq:eq2} of multi-layer fusion rendering. The composite rendering is defined as follows:

\begin{equation}
\begin{aligned}
& C_{comp}(r)=\sum_{x_i \in M_c} w_i c_c\left(x_i\right)+\sum_{x_j \in M_{bc}} w_j c_{bc}\left(x_j\right), \\
& M_c=\left\{x_i \mid w\left(x_i\right)<\lambda \cup \delta\left(x_i\right)>\delta\left(x_j\right)\right\}, \\
& M_{bc}=\bar{M}_c ,
\end{aligned}
\end{equation}
where $C_{comp}(.)$ is the rendering formula \cref{eq:eq2}, $x$ is the sampling point, $w(x)$ and $\delta(x)$ is the weight and density of $x$, $\lambda$ is the defined threshold, $c_{c}(x)$ and $M_{c}$ is the predicted color and set of $x$ for the $N^{th}$ cloth component, $c_{bc}(x)$ and $M_{bc}$ is the predicted color and set of $x$ for the first $N-1$ components.

\noindent 
\textbf{Dual SDS optimization.}
%这样我们会得到N个组件合成的渲染图像。但是直接使用该结果进行训练，会产生与服装语义不一致的问题。所以我们除了使用SDS损失对合成渲染结果进行监督外，我们还使用了一个单体积渲染结合Diffusion仅对服装进行监督。采用以上的解耦策略后，可以获得与身体解纠缠的服装。此外我们还引入了nerf密度正则化损失来消除浮云。解耦生成服装阶段的整体损失如下：
By using the above method, we obtain a rendered image composed of $N$ components. However, direct utilization of this outcome for the training of clothing leads to issues with semantic inconsistency with clothing. As shown in \cref{fig:pipline}(b), apart from utilizing SDS loss to supervise the composite rendering results, we also employ a single volume rendering combined with Diffusion model solely to supervise the clothing. After using the above decoupling strategy, we thus get clothing that is disentangled with the body. Additionally, we introduce NeRF density regularization loss $\mathcal{L}_r(.)$ to eliminate floating clouds. The loss function of the decoupling generation stage for clothing is as follows:
\begin{equation}
\mathcal{L}_{clothing}=\lambda_{\mathrm{SDS}}^{cloth} \mathcal{L}_{\mathrm{SDS}}^{cloth}+\lambda_{\mathrm{SDS}}^{comp} \mathcal{L}_{\mathrm{SDS}}^{comp}+\lambda_{reg\_ds} \mathcal{L}_{reg\_ds}+\lambda_r \mathcal{L}_r ,
\end{equation}
where $\mathcal{L}_\mathrm{SDS}^ {cloth}=\mathcal{L}_\mathrm{SDS}\left(\mathbf{x}^c; y^c\right)$, $\mathbf{x}^c$ is the supervised clothing image, $ y^c$ is the prompt of clothing.
$\mathcal{L}_\mathrm{SDS}^ {comp}=\mathcal{L}_\mathrm{SDS}\left(\mathbf{x}^{cp}; y^{cp}\right)$, $\mathbf{x}^{cp}$ is the supervised composite image, $ y^{cp}$ is the prompt of composite image.
$\lambda_{\mathrm{SDS}}^{cloth}$, $\lambda_{\mathrm{SDS}}^{comp}$, $\lambda_{reg\_ds}$, $\lambda_r$ are the weights attributed to each loss.

\subsection{Matching of Clothing and Body}
\label{sec:3.4}
In order to perform fine deformation of the clothing shape to fit the body, we introduce the SMPL-driven implicit deformation network (SID Net), as shown in \cref{fig:pipline}(b). Furthermore, for precise clothing editing, we use {SMPL-X}~\cite{a47} for our clothing shape proxy and add learnable vertex offsets $o$ for each shape proxy. At the same time, we use the vertex prediction model $o=F_v(v)$ to predict the offset $o$ of each vertex $v$ of the SMPL shape proxy. The specific implementation of SMPL to drive the clothing to match the body is as follows:

\noindent \textbf{Optimizing vertices.} Given the body SMPL parameters $({\beta}, {\theta})$, the vertex offset $F_v: \mathit{v} \rightarrow \mathit{o}$ and the camera parameter $\rho$, we render a mesh proxy of the body as a binary mask image $\mathcal{R}_m\left(M_{b o d y}({\beta}, {\theta}, \mathit{o}), \rho\right) \rightarrow I_{\mathit {smpl }}^{\mathit {body }}$, where $\mathcal{R}_m$ is a differentiable raster renderer. At the same time, we render a meshes proxy of the clothing as binary mask images ({where we use} the SMPL model {excluding} vertices of the head, hands and feet) $\mathcal{R}_m\left(M_{\mathit {cloth }}({\beta}, {\theta}), \rho\right) \rightarrow I_{\mathit {smpl }}^{\mathit {cloth }}$. Then, since the body masks should be within the region where the clothing proxy $I_{\mathit {mask }}^{\mathit {cloth }}$ rendered by NeRF and $I_{\mathit {smpl }}^{\mathit {cloth }}$ are merged, we perform the optimization using the following loss:
\begin{equation}
\begin{aligned}
\mathcal{L}_{\mathit {match }}=\mathcal{L}_{\mathit {huber }}\left(I_{\mathit {mask }}^{\mathit {cloth }}+I_{\mathit {smpl }}^{\mathit {cloth }}-I_{\mathit {smpl }}^{\mathit {body }}\right),
\end{aligned}
\end{equation}
where $\mathcal{L}_{\mathit {huber }}(.)$~\cite{a35} is a smoothed loss function. Also to smooth the predicted vertex offsets, we introduce a regularization loss for the vertex offset $o$:
\begin{equation}
\begin{aligned}
\mathcal{L}_{\mathit {offset }}^{\mathit {reg }}=\| \mathit { o } \|_2,
\end{aligned}
\end{equation}
where $o$ {contains} the predicted vertex offsets {for all vertices}. Then, we update the vertex prediction model $F_v: \mathit{v} \rightarrow \mathit{o}_{opt}$ using the gradient of the $\mathcal{L}_{match}$ loss to obtain the optimized vertex {offsets} $o_{opt}$ for optimizing the implicit geometry of the clothing. More details of \cref{sec:3.4} are provided in the supplementary material. The loss of clothing matching is delineated as follows:

\begin{equation}
%\vspace{-0.2cm}
%\begin{aligned}
\begin{split}  
\mathcal{L}_{matching}
&=\lambda_{\mathit {match }} \mathcal{L}_{\mathit {match }}+\lambda_{\mathit {reg }} \mathcal{L}_{\mathit {offset }}^{\mathit {reg }}
\end{split},
%\end{aligned}
\end{equation}
where $\lambda_{\mathit {match }}$, $\lambda_{\mathit {reg }}$ are the weights attributed to each loss.
In conclusion, the overall loss of the decoupled generation and matching of bodies and clothing is as follows:
\begin{equation}
%\vspace{-0.2cm}
\mathcal{L}_{all} = \mathcal{L}_{body }+ \mathcal{L}_{ {clothing }}+ \mathcal{L}_{ {matching }}.
\end{equation}

%Next, this paper presents experimental results.
%%%YKL redundant

%\vspace{-0.4cm}
\section{Experiments}
\label{sec:experiment}

In this section, we assess the efficacy of our proposed layered human generation framework. We commence by providing implementation details in~\cref{sec: imp_de}, followed by generated results in \cref{sec:generated_results}. Subsequently, quantitative and qualitative comparisons between state-of-the-art methods and ours are presented in~\cref{sec:5.1}. To evaluate the effectiveness of proposed modules, ablation studies are discussed in \cref{sec:5.7}. Finally, we showcase the applications of our method. Please refer to the demo case for some experimental results.

\subsection{Implementation Details}
\label{sec: imp_de}
\noindent \textbf{Hyperparameters.} We use ISM~\cite{a58} to compute the SDS loss with normal CFG ($7.5$) for all stages. The warm-up period of ISM is 1,000 iterations. (1) \emph{Canonical Body Generation:} The loss weights, $\lambda_{\mathrm{SDS}}^{\mathit{body}}$, $\!\lambda_n$ and $\!\lambda_n^{\mathit{reg}}$, are $1.0$, $0.01$, $0.05$, respectively. The gradient scaling factor of ISM is $0.1$. (2) \emph{Dual-Representation Decoupling:} The loss function weights, $\lambda_{\mathrm{SDS}}^{cloth}$, $\lambda_{\mathrm{SDS}}^{comp}$, $\lambda_{reg\_ds}$ and $\lambda_r$, are $1.0$, $1.0$, $0.05$, $2.0$, respectively. The gradient scaling factor of ISM is $0.07$. (3) \emph{Matching of Clothing and Body:} The loss weights, $\lambda_{\mathit {match}}$ and $\lambda_{\mathit {reg}}$, are $10.0$, $1.0$, respectively.

\noindent \textbf{Training Details.} The overall framework is trained using Adam optimizer, with the $betas$ of $[0.9, 0.99]$ and the learning rates of $5e-5$, $1e-3$ for the stage of decoupling dressed human and clothing-matching, respectively.
The training of the body and clothing in the decoupling stage takes 12,000 and 8,000 iterations. Specifically, alternate training is used in clothing training, and the training ratio of the $N$th layer to the combination of the first $N$ layers is $1:6$. 
The training of clothing-matching requires 3,000 iterations. We use the training resolution of $512 \times 512$ with a batch size of 2 and the whole optimization process takes three hours on a single NVIDIA 4096 GPU. Further training details are available in the SupMat.

% \begin{figure}[t]
%   \centering
%    \includegraphics[width=1\linewidth]{figures/C2.png}
%    %\vspace{-0.2cm}
%    \caption{User study. We investigated user evaluations on geometric and texture quality, as well as consistency with text prompts. }
%    \label{fig:C2}
%    %\vspace{-0.4cm}
% \end{figure}

\begin{figure}[!t]
  \centering   \includegraphics[width=1\linewidth]{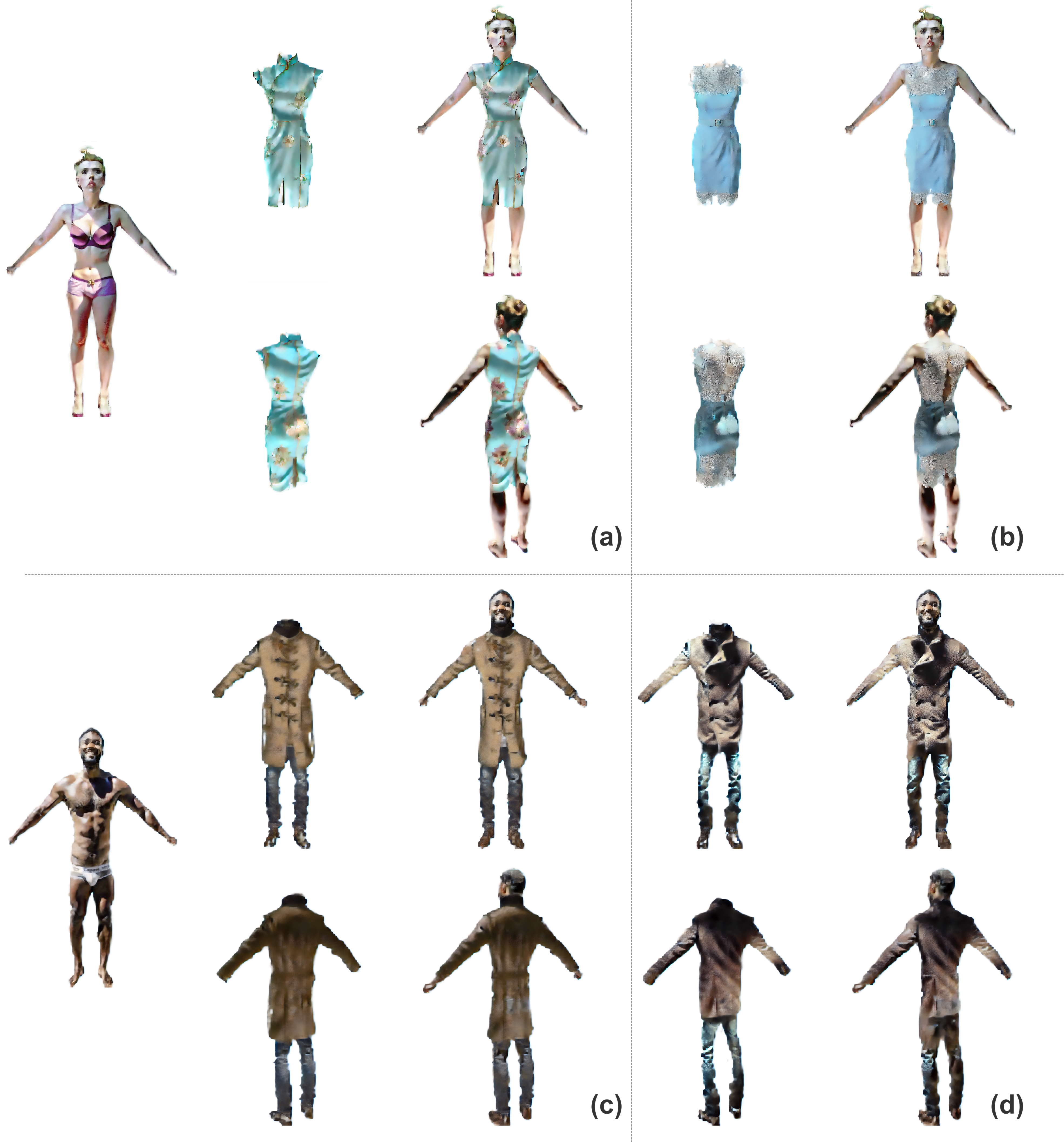}
   %\vspace{-0.2cm}
   \caption{The decoupled generation of human body and clothing by our method. (a) clothing prompt: ``A turquoise Cheongsam'', (b) clothing prompt: ``A deep-skyblue sleeveless sheath dress with lace trims'', (c) clothing prompt: ``A Duffle Coat and a baggy linen pants'', (d) clothing prompt: “A Car Coat and a baggy jeans”.}
   \label{fig:E5}
    %\vspace{-0.4cm}
\end{figure}

\begin{figure}[!h]
  \centering   \includegraphics[width=1\linewidth]{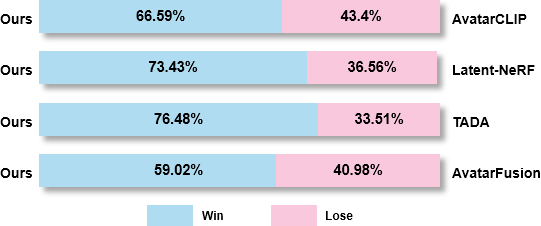}
   %\vspace{-0.2cm}
   \caption{Quantitative results. Our method and methods~\cite{a23,a11,a46,a54} are evaluated by using the method~\cite{a38} to measure the visual quality of the generated 3D content, where higher scores are better.}
   \label{fig:C1}
   %\vspace{-0.4cm}
\end{figure}

\begin{table}[!t]
  \caption{Quantitative comparisons with non-layered and Layered methods.}
  \centering
  \begin{tabular}{@{}lcc@{}}  
    \toprule
    Method & FID $\downarrow$ &CLIP Score $\uparrow$\\
    \midrule
    AvatarCLIP~\cite{a23}  &  311.46 & 30.88\\
    Latent-NeRF~\cite{a11}  & 329.40 & 29.82\\
    TADA~\cite{a46}  & 392.61 &  25.39\\
    \midrule
    AvatarFusion\cite{a54}   & 375.97 &26.96 \\
    HumanLiff~\cite{a48}  &  324.69  & 26.34 \\
    \midrule
    HumanCoser (Ours) &  {\color[rgb]{0.19,0.4,0.67} \textbf{298.54}}  &  {\color[rgb]{0.19,0.4,0.67} \textbf{31.61}} \\
    \bottomrule
  \end{tabular}
    %\vspace{-5pt}
  \label{tab:fid_clip_score}
  %\vspace{-10pt}
\end{table}

\begin{table*}[!t]
\caption{User Study Results. We investigated user evaluations on geometric and texture quality, as well as consistency with text prompts.}
\centering
%\scalebox{1}{
\setlength{\tabcolsep}{2mm}{
\begin{tabular}{|c|ccc|ccc|ccc|ccc|}
\hline
& \multicolumn{3}{c|}{\textbf{AvatarCLIP}~\cite{a23}} 
& \multicolumn{3}{c|}{\textbf{TADA}~\cite{a46}}
& \multicolumn{3}{c|}{\textbf{Latent-NeRF}~\cite{a11}}
& \multicolumn{3}{c|}{\textbf{Ours}} \\ \cline{2-13} 
\centering{Case} & Geometry  & Texture   & Text & Geometry  & Texture  & Text  & Geometry & Texture   & Text & Geometry & Texture & Text            \\ \hline
\textbf{case 1}                       & 2.41                                 & 2.82                                 & 2.88                                 & 2.58                                 & 2.74                                 & 2.34                                 & 3.18                                 & 3.22                                 & 4.04                                 & 3.72                                 & 4.26                                 & 4.29                                 \\ \hline
\textbf{case 2}                       & 3.85                                 & 2.79                                 & 2.85                                 & 2.24                                 & 2.48                                 & 2.72                                 & 4.06                                 & 2.86                                 & 3.29                                 & 4.53                                 & 4.51                                 & 4.16                                 \\ \hline
\textbf{case 3}                       & 3.05                                 & 3.07                                 & 2.33                                 & 2.47                                 & 2.31                                 & 2.42                                 & 3.58                                 & 2.27                                 & 3.82                                 & 4.61                                 & 3.79                                 & 4.63                                 \\ \hline
\textbf{case 4}                       & 2.57                                 & 2.51                                 & 3.27                                 & 2.78                                 & 2.43                                 & 2.51                                 & 3.12                                 & 3.76                                 & 2.89                                 & 3.08                                 & 3.74                                 & 4.53                                 \\ \hline
\textbf{case 5}                       & 3.24                                 & 2.46                                 & 2.74                                 & 2.54                                 & 2.10                                  & 2.03                                 & 3.41                                 & 3.54                                 & 3.24                                 & 4.66                                 & 3.64                                 & 3.79                                 \\ \hline
\textbf{case 6}                       & 2.59                                 & 2.41                                 & 3.95                                 & 3.02                                 & 2.61                                 & 2.49                                 & 3.14                                 & 3.57                                 & 3.17                                 & 4.16                                 & 3.93                                 & 4.89                                 \\ \hline
\textbf{case 7}                       & 2.37                                 & 2.60                                  & 2.58                                 & 2.67                                 & 1.83                                 & 2.17                                 & 3.70                                  & 3.43                                 & 3.95                                 & 4.68                                 & 4.13                                 & 4.23                                 \\ \hline
\textbf{case 8}                       & 2.55                                 & 3.11                                 & 2.08                                 & 2.57                                 & 2.26                                 & 1.90                                  & 2.97                                 & 3.65                                 & 3.71                                 & 4.32                                 & 3.73                                 & 4.64                                 \\ \hline
\textbf{case 9}                       & 2.88                                 & 2.93                                 & 3.08                                 & 2.51                                 & 2.10                                  & 2.33                                 & 3.87                                 & 3.81                                 & 3.22                                 & 4.47                                 & 4.40                                  & 4.11                                 \\ \hline
\textbf{case 10}                      & 3.79                                 & 2.40                                  & 2.64                                 & 1.92                                 & 2.74                                 & 2.29                                 & 2.47                                 & 2.79                                 & 2.97                                 & 4.37                                 & 3.97                                 & 4.36                                 \\ \hline

\textbf{Average}  & 3.79 & 2.71 & 2.84 & 2.53 & 2.36 & 2.32 & 3.35 & 3.29 & 3.43 & {\color[rgb]{0.19,0.4,0.67} \textbf{4.26}}  & {\color[rgb]{0.19,0.4,0.67} \textbf{4.01}}  & {\color[rgb]{0.19,0.4,0.67} \textbf{4.37}}  \\ \hline
% \multicolumn{1}{l|}{\textbf{Average}} & 3.79 & 2.71 & 2.84 & 2.53 & 2.36 & 2.32 & 3.35 & 3.29 & 3.43 & \textbf{4.26} & \textbf{4.01} & \textbf{4.37} \\ \hline
\end{tabular}
}
\label{tab:userstudy}
\end{table*}

\begin{figure}[!b]
  \centering
   \includegraphics[width=1\linewidth]{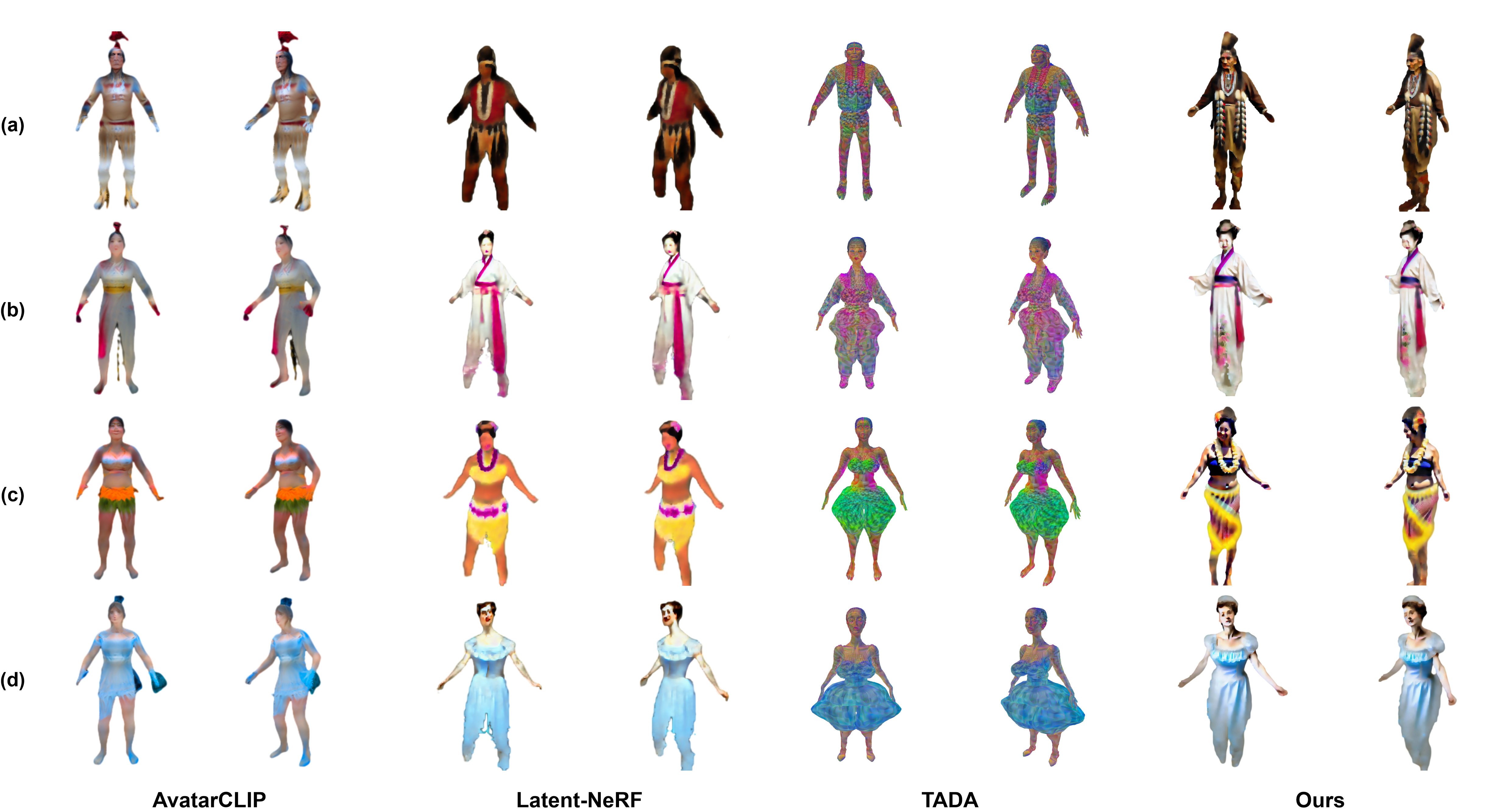}
   %\vspace{-0.2cm}
   \caption{Qualitative comparison with coupled generation methods~\cite{a23,a46,a1}. (a) prompt: ``A north {American Indian} chief in full regalia'', (b) prompt: ``A Chinese lady wearing a gauzy hanfu'', (c) prompt: ``A Hawaiian woman wearing a hula skirt'', (d) prompt: “A French woman wearing a light blue crinoline dress”.}
   \label{fig:B1}
    %\vspace{-0.4cm}
\end{figure}

\subsection{Generated Results}
\label{sec:generated_results}
We present physically-layered generated results in \cref{fig:E5}. When the same person is dressed in different clothes, our method generates 3D clothing that conform to the body shape. For example, when a woman wears a turquoise Cheongsam and then switches to a blue dress, our method generates her in the dress with a fitting body shape. In addition to all-in-one clothing, our method is capable of generating 3D humans in complex clothes, such as a man wearing a coat and pants or jeans. Notably, the generated clothes conform better to the body, including the waist position, suggesting that our physically-layered model not only accommodates various clothing changes but also ensures a better fit to the human body, resulting in a more natural appearance.

%\vspace{-0.2cm}
\subsection{Comparison}
\label{sec:5.1}
%% \vspace{-0.1cm}
We compare our approach with {five} SoTA methods. (1) AvatarCLIP~\cite{a23} uses pre-trained vision-language CLIP model to guide {NeuS}~\cite{a37} for 3D avatar generation; (2) TADA~\cite{a46} creates 3D avatars from text by using hierarchical rendering with score distillation sampling; (3) Latent-NeRF~\cite{a11} introduces sketch shape loss based on 3D shape guidance to supervise the training; (4) AvatarFusion~\cite{a54} can generate avatars while simultaneously segmenting clothing from the avatar's body; (5) HumanLiff~\cite{a48} firstly generates minimally clothed humans, represented by tri-plane features, in a canonical space and then progressively generates clothes in a layer-wise manner.

\subsubsection{Quantitative Results}
\label{sec:5.3}
This section quantitatively compares the proposed method with~\cite{a23,a46,a11,a54,a48}. Inspired by~\cite{a38}, we use user preference metrics to compare the generation quality to the SoTA methods~\cite{a23,a46,a11,a54}. ~\cref{fig:C1} {demonstrates} the superior performance of our method compared to~\cite{a23, a46, a11,a54} in generation quality. Additionally, we calculate the FID~\cite{a56} between the views rendered from the generated 3D humans and the images produced by Stable Diffusion~\cite{a57}. As shown in the \cref{tab:fid_clip_score}, our method achieves the lowest FID score, indicating the best generation quality. Furthermore, we adopt CLIP score~\cite{a55} to measure the compatibility between the prompts with the rendered views of 3D humans. \cref{tab:fid_clip_score} shows our method achieves the highest CLIP score, indicating that the human model generated by our framework is more aligned with the prompt. Compared to layered-generation SoTA methods~\cite{a54} and~\cite{a48}, our method not only achieves better generation quality, but also freely performs clothing transfer and generalizable animations.

\iffalse
In addition, {we} perform a user study comparing our avatar generation with those of SoTA methods~\cite{a23,a46,a11}, as shown in \cref{fig:C2}. 
Our approach achieves {best} scores across all three metrics, suggesting superior generative quality in both geometry and texture.
\fi

Furthermore, we perform a user study comparing the human generation results of our method with those of other state-of-the-art methods~\cite{a23,a46,a11}. We generate 3D human for different methods based on 10 text prompts. Fifty volunteers (including 26 males and 24 females, aged between 18 and 50 years) were invited to rank the methods in terms of (1) geometric quality, (2) appearance quality, and (3) consistency with the text prompts. Volunteers score each comparative indicator for each method from 1 (worst) to 5 (best). The final evaluation results are provided in \cref{tab:userstudy}. Our method achieves optimal scores across all three metrics, indicating superior generative quality for geometry and texture based on text inputs.

\begin{figure}[t!h]
  \centering
   \includegraphics[width=1.0\linewidth]{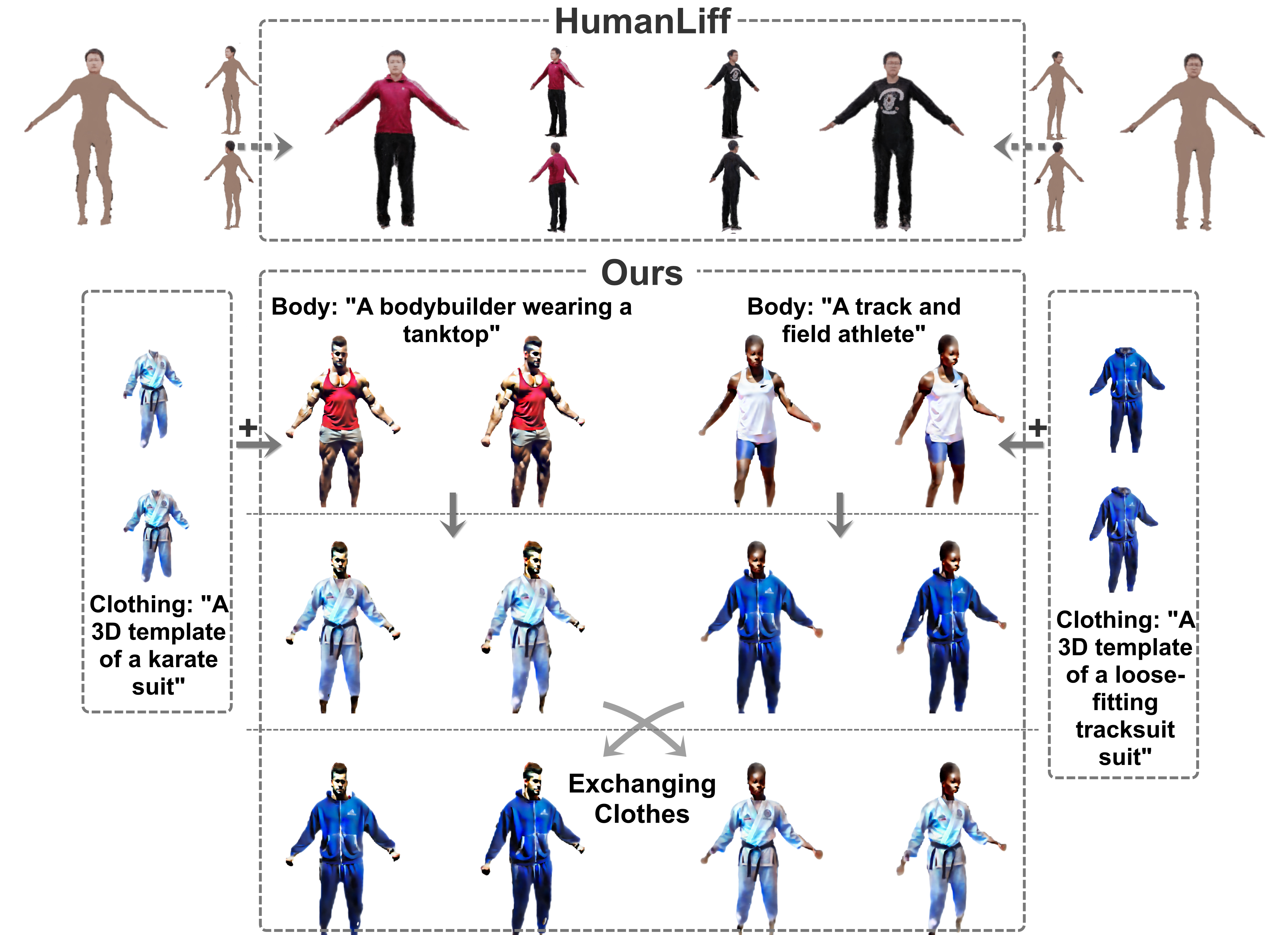}
   %\vspace{-0.2cm}
   \caption{Qualitative comparison with the layered method~\cite{a48}.}
   \label{fig:B5}
    %\vspace{-0.4cm}
\end{figure}

\begin{figure}[t!h]
  \centering
   \includegraphics[width=1.0\linewidth]{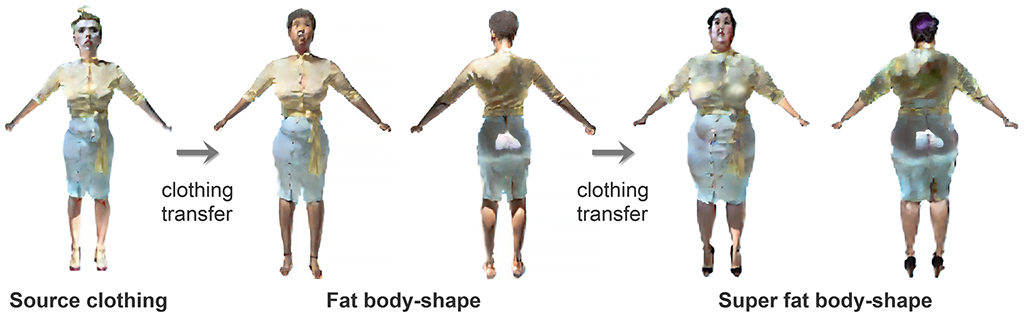}
   %\vspace{-0.2cm}
   \caption{Editing results for adaptive matching of clothing to different body shapes.}
   \label{fig:E6}
    %\vspace{-0.4cm}
\end{figure}

\begin{figure*}[t!h]
  \centering
   \includegraphics[width=0.9\linewidth]{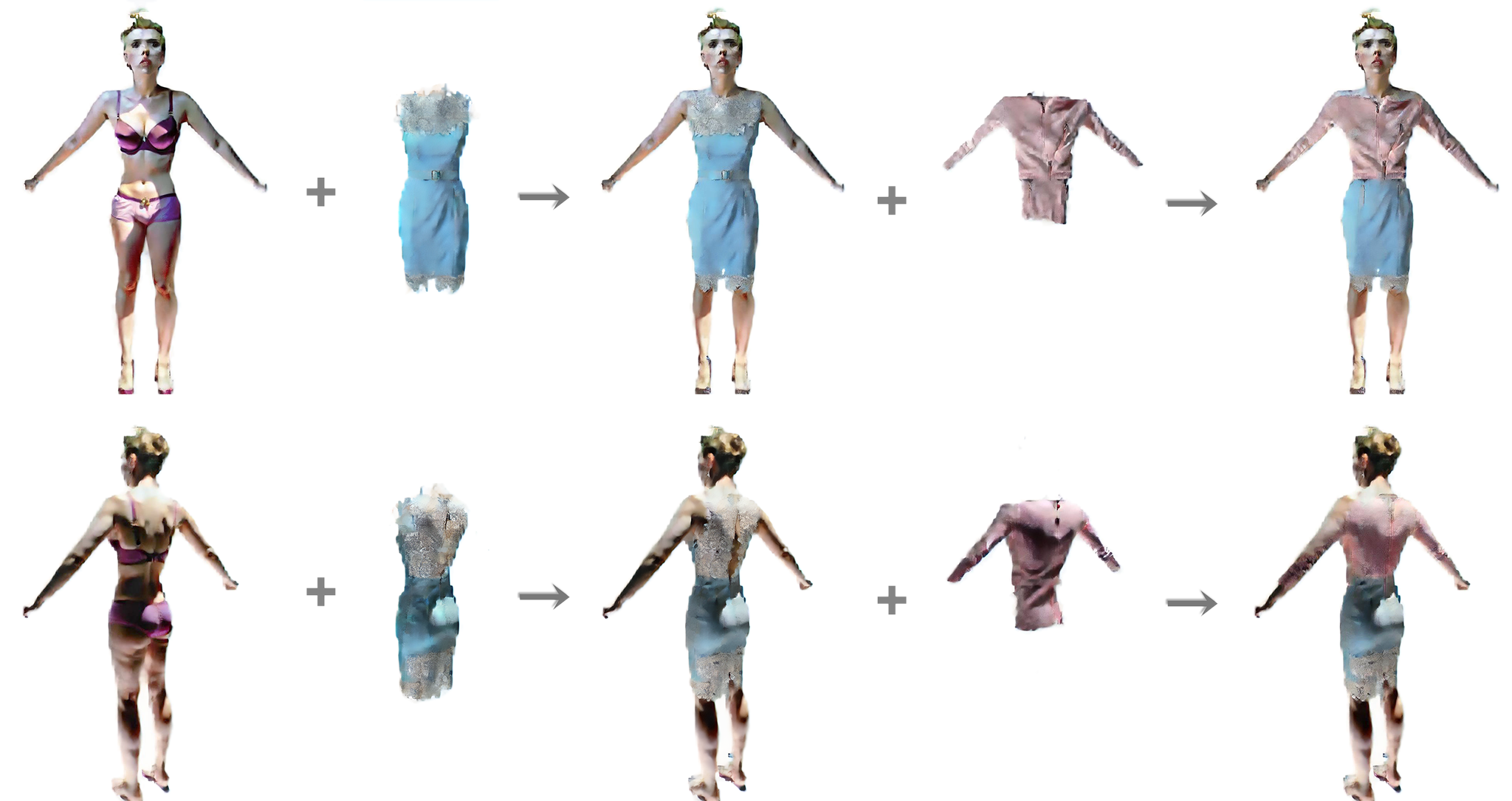}
   %\vspace{-0.2cm}
   \caption{The effectiveness of multi-layer decoupled clothing. }
   \label{fig:E4}
   %\vspace{-0.4cm}
\end{figure*}

\begin{figure}[t!h]
  \centering   \includegraphics[width=1.0\linewidth]{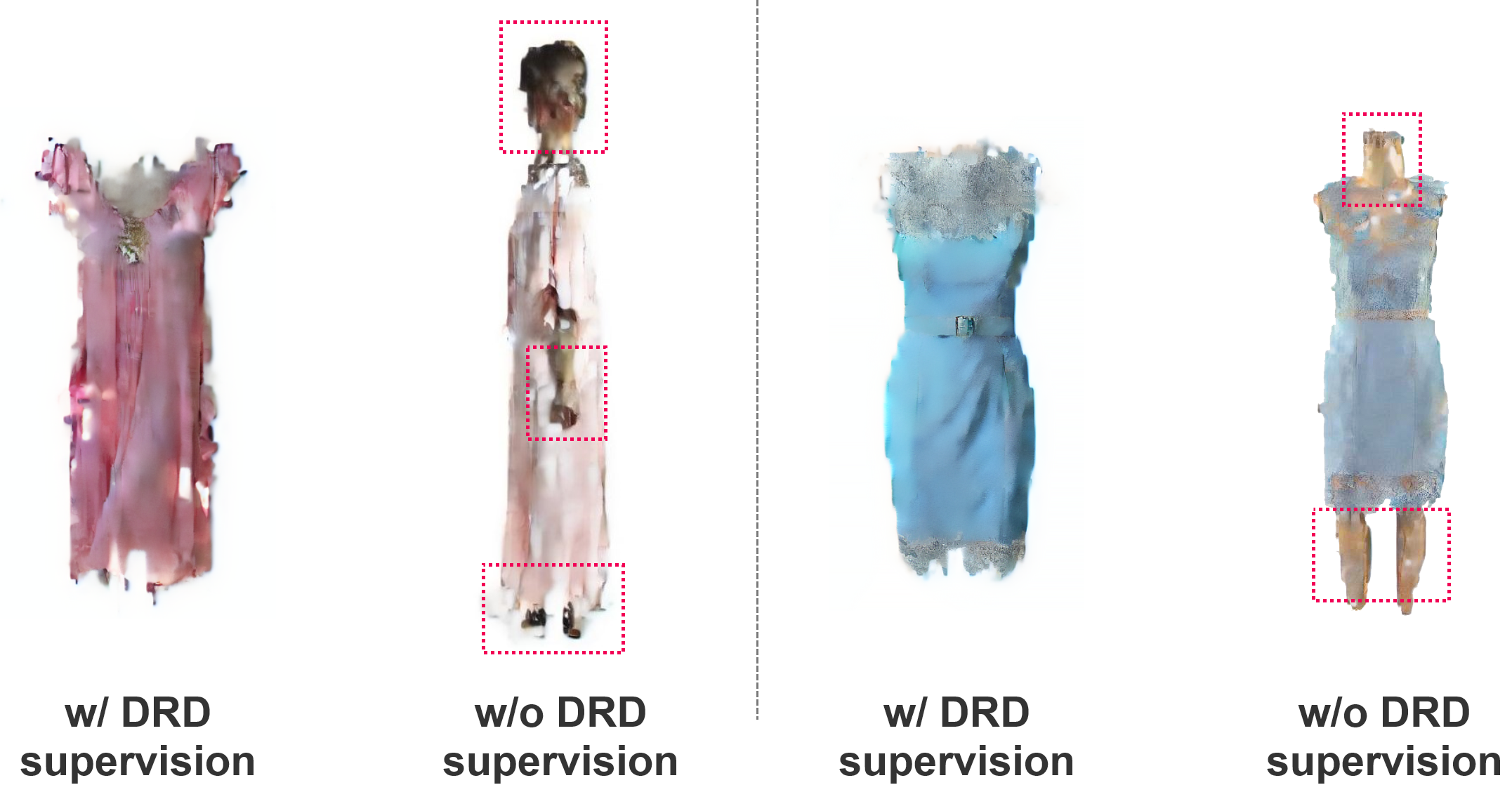}
   %\vspace{-0.2cm}
   \caption{Ablation study on the effectiveness of the dual-representation decoupling framework.}
   \label{fig:E1}
   %\vspace{-0.4cm}
\end{figure}

\begin{figure}[t!h]
  \centering   \includegraphics[width=1\linewidth]{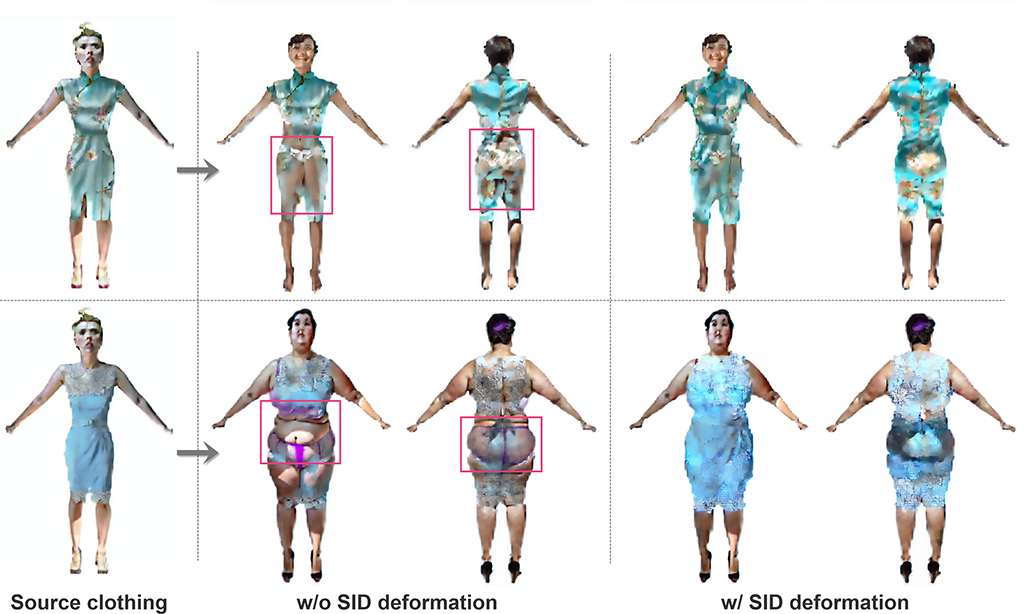}
  %\vspace{-0.2cm}
   \caption{Ablation study on the implicitly deformed modules.}
   \label{fig:deform}
   %\vspace{-0.4cm}
\end{figure}
%\vspace{-0.2cm}

\begin{figure*}[t!]
  \centering
\includegraphics[width=0.9\linewidth]{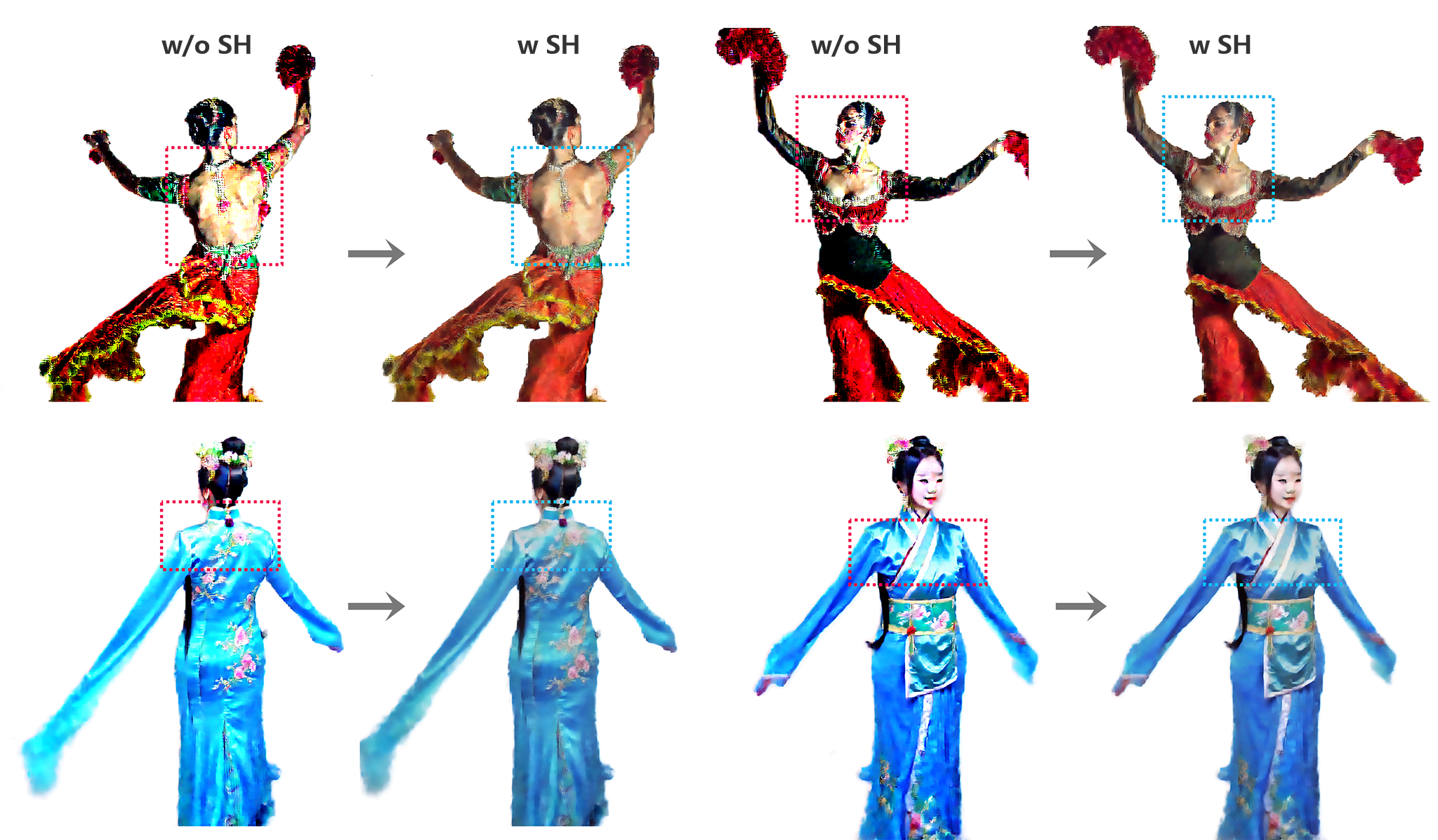}
  %\vspace{-0.2cm}
   \caption{Ablation study on the effectiveness of spherical harmonic (SH) lighting.}
   \label{fig:E2}
   %\vspace{-0.4cm}
\end{figure*}

%\vspace{-0.2cm}
\subsubsection{Qualitative Results}
\label{sec:5.2}
\cref{fig:B1} qualitatively compares to text-guided 3D generation methods~\cite{a23,a11,a46}. Considering that~\cite{a23,a11,a46} are based on coupled generation, we provide a coupled generation model for comparison. We render the model as multiple views for comparison. As shown in \cref{fig:B1}, although AvatarCLIP~\cite{a23} generates view-consistent human bodies, it demonstrates limitations in effectively modeling global structures, such as skirts and long hair. Latent-NeRF~\cite{a11} exhibits a limitation in its capacity to finely generate both geometry and texture. TADA~\cite{a46} accuracy depends on the density of the mesh, and the discrete representation affects its geometric appearance. So, ~\cite{a23,a11,a46} exhibit deficiencies, either in the representation of geometric details or in the portrayal of fine textures. In contrast, our method produces humans characterized by enhanced geometric details, including loose clothing and diverse long hair, along with finer textures.

In addition, \cref{fig:B5} illustrates the comparison of the layered 3D human generation approaches~\cite{a48}. Since AvatarFusion~\cite{a54} is not capable of  multi-layer generation, we use HumanLiff~\cite{a48}\footnote{HumanLiff currently does not provide the official implementation, and hence we compare with the visual results presented in~\cite{a48}} for the comparison of layered generation. HumanLiff~\cite{a48} stands out as the most akin work to our method, employing a layer-by-layer generation approach. However, it lacks the capability to change clothes, as illustrated in the top row. HumanLiff generates a clothed human body by relying on a {minimally-clothed} human body. Instead, our method demonstrates the ability to generate the body and clothing independently, as depicted in the second row in \cref{fig:B5}. Subsequently, it engages in the matching of clothing and body, showcased in the third row. Finally, our method excels in the process of changing and reusing clothing, as illustrated in the last row in \cref{fig:B5}.

It's important to highlight that our method not only facilitates the transfer and matching of clothing across bodies of varying shapes but also enables the generation of multi-layer clothing using multi-layer fusion volume rendering. \cref{fig:E6} shows that the clothing can adaptively match different shapes of body by our method including even extreme body shapes, i.e. the ``super fat woman''. \cref{fig:E4} shows that a lady is wearing two-layer clothes, i.e. a dress as well as an outer clothing. Two distinct views showcase the harmony and naturalness achieved by our method in multi-layer clothing.

%\vspace{-0.0cm}
\subsection{Ablation Study}
\label{sec:5.7}

\iffalse
\noindent \textbf{Effectiveness of Layered Shape Prior.} To verify the effectiveness of the layered shape prior, we generate humans with or without the shape prior.
{Figs.~\ref{fig:D3} (a, b) show} that when we remove the layered shape prior, the model seems not to converge correctly to a reasonable human body shape and pose. This is due to the complexity of the human body structure and self-occlusion. Additionally, the elimination of the layered shape prior during the clothing generation process leads to either an inability to converge to a well-defined geometry, as depicted in \cref{fig:D3} (c), or suboptimal convergence in both geometry and texture, exemplified in \cref{fig:D3} (d). Hence, the layered shape prior ensures that the model converges to the correct shape.
\fi

\noindent 
\textbf{Effectiveness of Dual-Representation Decoupling Framework.} 
% To verify the effectiveness of the dual-representation decoupling framework, we analyze the effect of the dual SDS losses on clothing generation, as shown in \cref{fig:E1}. The results show that when only using a single SDS loss and a single volumetric rendering, the clothing will not be correctly decoupled from the human body, and even cannot generate the correct clothing shape. Because a single SDS loss supervises clothing generation, it will produce redundant non-clothing parts. Instead, when adding additional SDS to supervise the combination results of the human body and the clothing, the redundant non-clothing parts will be eliminated and the semantic consistency with the clothing will be maintained. Therefore, the proposed dual-representation decoupling framework is effective for generating intricate and semantically consistent clothing.
To assess the effectiveness of the dual-representation decoupling framework (DRD), we investigate the impact of employing dual SDS losses on clothing generation, as depicted in \cref{fig:E1}. Our findings indicate only utilizing a single SDS loss alongside a single volumetric rendering fails to accurately decouple the clothing from the human body and may result in incorrect clothing shapes as shown in the red box in \cref{fig:E1}. This is attributed to the single SDS loss supervising clothing generation, leading to the production of redundant non-clothing parts. However, by incorporating additional SDS to supervise the combined results of the human body and the clothing, we observe a significant improvement. This augmentation enables the elimination of redundant non-clothing parts and maintains semantic consistency with the clothing. Consequently, the proposed dual-representation decoupling framework validates its efficacy in generating intricate and semantically consistent clothing.

\noindent 
\textbf{Effectiveness of Implicitly Deformed Modules.}
To adaptively match the decoupled clothing to different body shapes, we introduce the SMPL-driven implicit field deformation network (SID Net). As seen from the red boxes in \cref{fig:deform}, the decoupled clothing is directly matched to different body shapes, which leads to the issue of interpenetration between the clothing and the body, and the clothing does not fit tightly and naturally to the body. Our SID Net can optimize the SMPL proxy model of the clothing to deform the implicit field of the clothing to match the body by calculating the shape deviation loss between the clothing and the body. As can be seen from columns 4,5 of \cref{fig:deform}, arbitrarily decoupled clothing can be freely and accurately matched with bodies of different shapes, even including extreme shapes of the human body, such as a super-fat or a very thin person. Our SID Net is validated to efficiently perform adaptive clothing-body matching via the above visualization results.

\noindent 
\textbf{Effectiveness of Optimizable Spherical Harmonic (SH) Lighting.} 
% As described in \cref{sec:3.2}, to address the issue of color oversaturation caused by SDS loss in diffusion model, we added an optimizable spherical lighting to the color of the sample point. As shown in the red box in \cref{fig:E2}, the color of the 3D dressed human without adding spherical lighting produces oversaturation and unsmooth surface. The blue box in \cref{fig:E2} shows that the human model can obtain the correct color and smooth effect after adding spherical lighting.
As detailed in \cref{sec:3.2}, to mitigate the problem of color oversaturation stemming from SDS loss in the diffusion model, we introduced an optimizable SH lighting component to modulate the color of the sample point. As depicted in the red box in \cref{fig:E2} without incorporating SH lighting, the color of the 3D dressed human exhibits oversaturation and lacks smoothness in surface rendering. Contrastingly, the blue box in \cref{fig:E2} illustrates that integrating SH lighting enables the human model to achieve the correct coloration and a smoother visual effect. This enhancement not only addresses the issue of oversaturation but also contributes to improving the overall realism and visual fidelity of the rendered human models. The addition of SH lighting introduces subtle variations in color and shading, resulting in a more natural appearance that better aligns with real-world lighting conditions. Hence, this approach enhances the quality and believability of the generated results, providing more accurate representations of dressed human subjects.

\subsection{Application}
\label{sec:application}
Thanks to our capability of generating layered 3D humans, our method also has the ability to transfer clothing across people and enable skeleton-driven layered human animation.

\begin{figure}[t!]
  \centering
\includegraphics[width=1\linewidth]{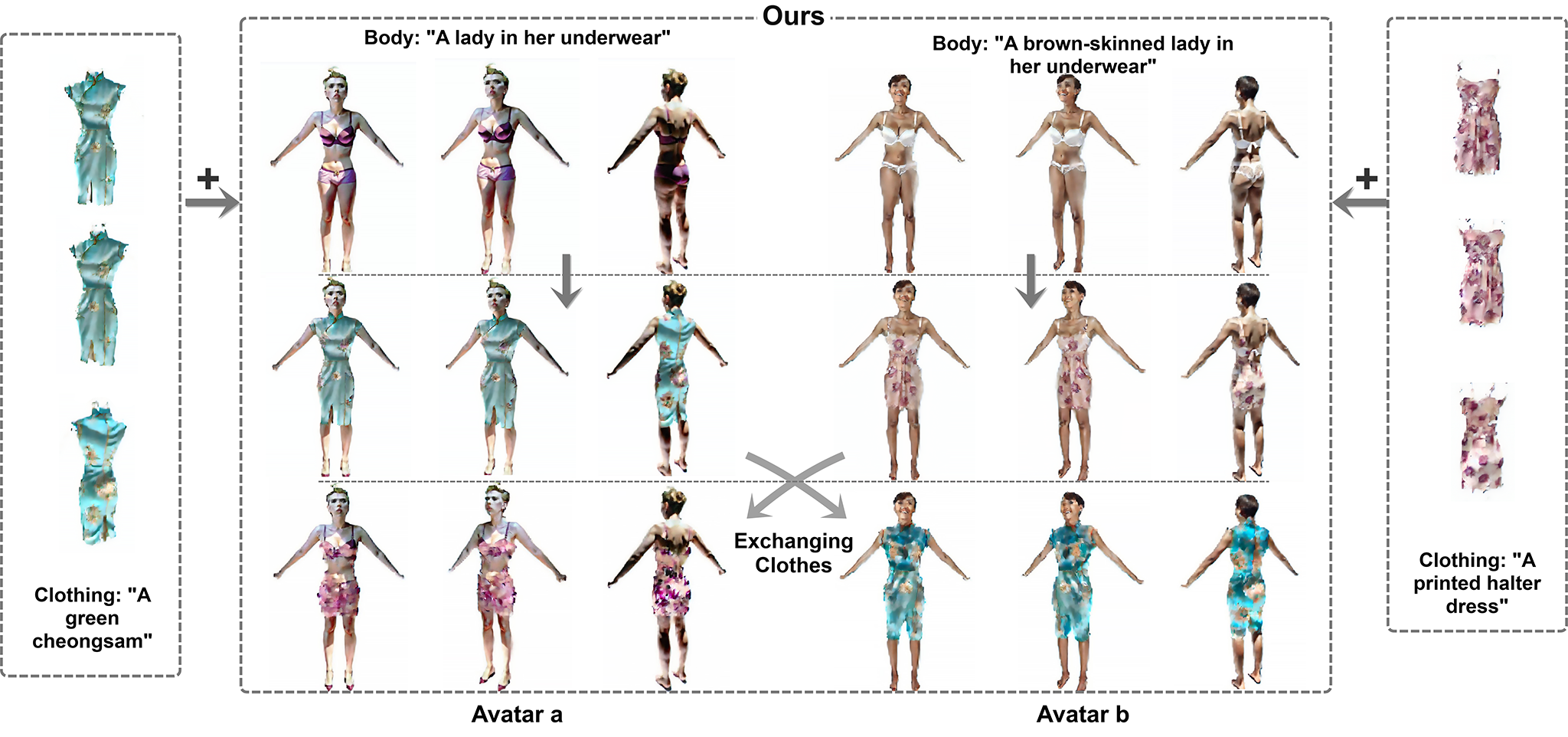}
   %\vspace{-0.2cm}
   \caption{The effectiveness of clothing transfers.}
   \label{fig:B2}
   %\vspace{-0.4cm}
\end{figure}

\noindent\textbf{Clothing Transfer.} 
\cref{fig:B2} evaluates the effectiveness of our model in clothing transfer by exchanging avatars' clothes (left/right). In this case, the layered avatars are generated based on different SMPL {shapes} $\theta$ with the same pose $\beta$. 
We transfer the clothing layer of avatar left to the body layer of avatar right {and vice versa}: $\left(\right.$cloth ${ }^{\text{avatar left}}\rightarrow$ body${}^{\text{avatar right}}$, cloth$^{\text{avatar right }}\rightarrow$body$\left.{ }^{\text{avatar left}}\right)$. \cref{fig:B2} illustrates our model excels in adaptively shaping a match between the body and clothing layers, facilitating the transfer of the same clothing layer across different identity-based body layers.

\begin{figure}[t!]
  \centering
   \includegraphics[width=0.88\linewidth]{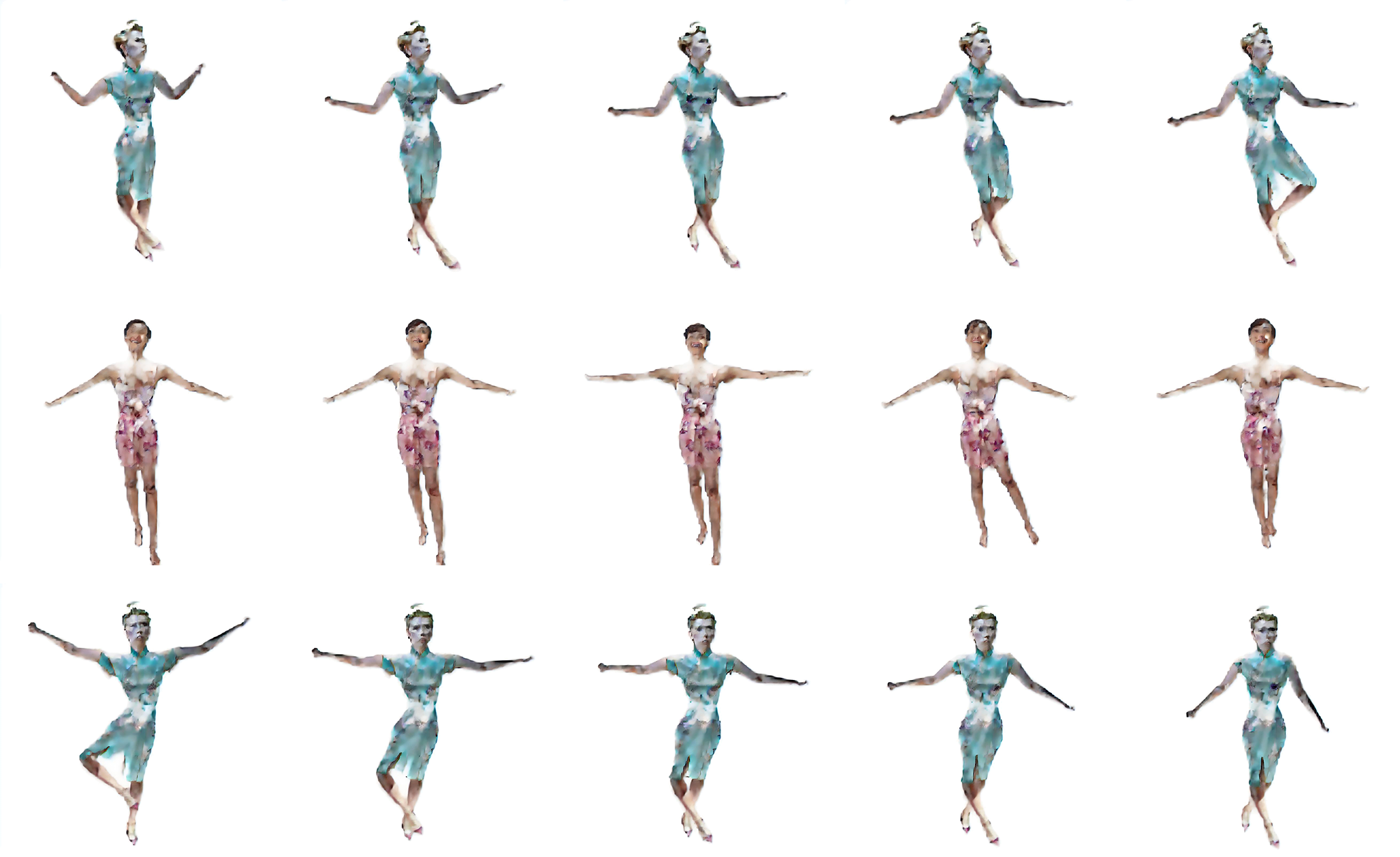}
   % \vspace{-0.2cm}
   \caption{The effectiveness of Pose-Driven Generation.}
   \label{fig:B3}
   % \vspace{-0.4cm}
\end{figure}

\noindent\textbf{Generalizable Poses and Animations.} 
\cref{fig:B3} demonstrates the effectiveness of SMPL skeleton-driven layered human animation by applying complex animations and poses to the body and clothing layers. We learn a generalizable density-weighted network by sampling the pose of the SMPL from the pre-trained VPoser model as conditional inputs to the ControlNet. This refines the SMPL-based pose deformations and supports SMPL-driven animations and complex poses without additional training.

\section{Conclusion and Limitations}
\label{sec:conclusion}
\noindent\textbf{Conclusion.} 
This paper introduces a layer-wise dressed human generation framework built upon a physically-decoupled diffusion model. Central to our approach are the concepts of a dual-representation decoupling framework and a novel multi-layer fusion volumetric rendering technique. 
Building upon this decoupled representation, we achieve multi-layer 3D human wearing loose-fitting clothing while the existing coupled methods struggle to achieve layered dressed human. Additionally, unlike other methods that fail to arbitrarily change and exchange clothing, we introduce an implicit deformation module, guided by the SMPL model, which allows clothing to adaptively match different body shapes. Experimental results showcase that our method outperforms state-of-the-art approaches by generating high-quality multi-layered 3D humans wearing complex clothing and arbitrarily switching clothing across various body shapes.

\noindent\textbf{Limitations.} 
Given the absence of a uniform parametric clothing template, the assessment of matching loss to the body cannot be conducted through differentiable rendering employing a uniform 3D proxy tailored to the generated clothing. Consequently, we opt for a 3D implicit deformation field based on {SMPL-X}~\cite{a47} to optimize the alignment between bodies and clothing. While our method enables the fitting of the clothing to various body shapes, it may yield unnatural matching outcomes when the shapes of the body and clothing differ significantly. In future, we will employ more accurate deformation proxies combined with object collision detection to optimize the matching of clothing and body bidirectionally in order to achieve better quality of layered generation.

\acknowledgments{
This work was supported in part by the National Natural Science Foundation of China (62122058 and 62171317), and the Science Fund for Distinguished Young Scholars of Tianjin (No. 22JCJQJC00040).}

%%%%%%%%%%%%%hero end%%%%%%%%%%%%%%

%\bibliographystyle{abbrv}
\bibliographystyle{abbrv-doi}

\bibliography{template}
\end{document}